\documentclass[10pt,twocolumn,letterpaper]{article}

\usepackage{amsmath,amsfonts,bm}

\def\eqref#1{equation~\ref{#1}}

\def\1{\bm{1}}

\def\vp{{\bm{p}}}

\def\vw{{\bm{w}}}

\def\evp{{p}}

\def\evw{{w}}

\def\mM{{\bm{M}}}

\def\mW{{\bm{W}}}

\DeclareMathAlphabet{\mathsfit}{\encodingdefault}{\sfdefault}{m}{sl}
\SetMathAlphabet{\mathsfit}{bold}{\encodingdefault}{\sfdefault}{bx}{n}

\def\emM{{M}}

\usepackage{cvpr}
\usepackage{times}
\usepackage{epsfig}
\usepackage{graphicx}
\usepackage{amsmath}
\usepackage{amssymb}
\usepackage{graphicx}
\usepackage[titlenumbered,ruled]{algorithm2e}
\usepackage{algorithmic}
\usepackage{multirow}
\usepackage{multicol}
\usepackage{makecell}
\usepackage{subcaption}
\usepackage{eqparbox}
\usepackage{mathtools}
\usepackage{nccmath}
\usepackage{enumitem}
\newcommand{\mr}[2]{\multirow{#1}{*}{\begin{tabular}[c]{@{}c@{}}#2\end{tabular}}}

\usepackage[pagebackref=true,breaklinks=true,letterpaper=true,colorlinks,bookmarks=false]{hyperref}

\cvprfinalcopy 

\def\cvprPaperID{2609} 

\ifcvprfinal\fi
\begin{document}

\title{Structured Compression by Weight Encryption \\for Unstructured Pruning and Quantization}

\author{Se Jung Kwon\textsuperscript{1}\thanks{Equal contribution} \qquad Dongsoo Lee\textsuperscript{1}\footnotemark[1] \qquad Byeongwook Kim\textsuperscript{1} \\Parichay Kapoor\textsuperscript{1} \qquad Baeseong Park\textsuperscript{1} \qquad Gu-Yeon Wei\textsuperscript{1,2}\\
\textsuperscript{1}Samsung Research, Republic of Korea \hspace{30pt}\textsuperscript{2}Harvard University, MA\\
{\tt\small \{sejung0.kwon, dongsoo3.lee, byeonguk.kim, pk.kapoor, bpbs.park, gy.wei\}@samsung.com}
}

\maketitle

\begin{abstract}
  Model compression techniques, such as pruning and quantization, are becoming increasingly important to reduce the memory footprints and the amount of computations.
  Despite model size reduction, achieving performance enhancement on devices is, however, still challenging mainly due to the irregular representations of sparse matrix formats.
  This paper proposes a new weight representation scheme for Sparse Quantized Neural Networks, specifically achieved by fine-grained and unstructured pruning method.
  The representation is encrypted in a structured regular format, which can be efficiently decoded through XOR-gate network during inference in a parallel manner.
  We demonstrate various deep learning models that can be compressed and represented by our proposed format with fixed and high compression ratio.
  For example, for fully-connected layers of AlexNet on ImageNet dataset, we can represent the sparse weights by only 0.28 bits/weight for 1-bit quantization and 91\% pruning rate with a fixed decoding rate and full memory bandwidth usage.
  Decoding through XOR-gate network can be performed without any model accuracy degradation with additional patch data associated with small overhead.
\end{abstract}

\section{Introduction}

Deep neural networks (DNNs) are evolving to solve increasingly complex and varied tasks with dramatically growing data size \cite{deeplearningbook}.
As a result, the growth rate of model sizes for recent DNNs leads to slower response times and higher power consumption during inference \cite{EIE}.
To mitigate such concerns, model compression techniques have been introduced to significantly reduce model size of DNNs while maintaining reasonable model accuracy.

\begin{figure*}[th]
\begin{center}
\centerline{\includegraphics[width=0.85\linewidth]{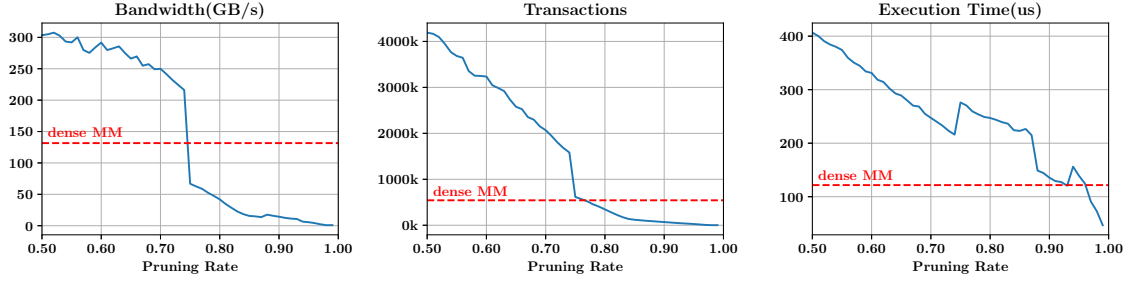}}
\caption{DRAM bandwidth, the number of transactions, and execution time of a matrix multiplication using one random $(2048 \times 2048)$ sparse matrix (following CSR format) and a random $(2048 \times 64)$ dense matrix using NVIDIA Tesla V100. CUDA 9.1 is used as a main computation library and analysis is supported by NVIDIA Profiler. Performance of a multiplication using two dense matrices (denoted as dense MM) without pruning is also provided as a baseline.}
\label{intro:cuda}
\end{center}
\end{figure*}

\begin{figure}[]
\begin{center}
\centerline{\includegraphics[width=0.9\columnwidth]{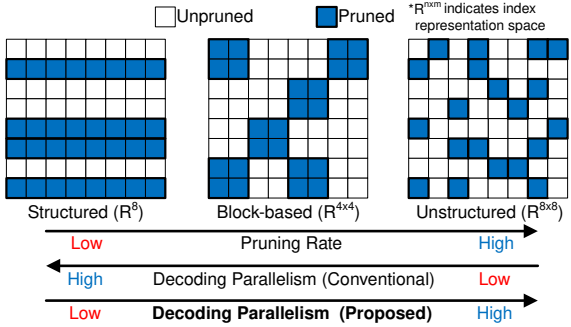}}
\caption{Several types of pruning granularity. In the conventional sparse formats, as a sparse matrix becomes more structured to gain parallelism in decoding, pruning rate becomes lower in general.}
\label{intro:pi}
\end{center}
\end{figure}

\begin{figure}
\begin{center}
\centerline{\includegraphics[width=0.9\columnwidth]{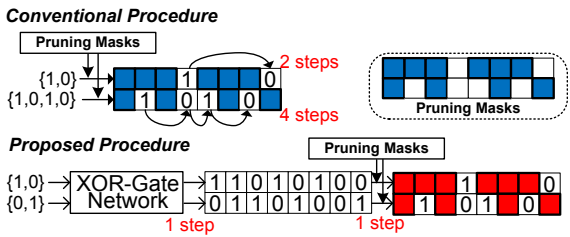}}
\caption{Comparison between conventional and proposed sparse matrix decoding procedures given a pruning mask. In the conventional approach, the number of decoding steps for each row can be different (i.e., degraded row-wise parallelism). On the contrary, the proposed approach decodes each row at one step by using XOR-gate network.}
\label{intro:recon}
\end{center}
\end{figure}
It is well known that DNNs are designed to be over-parameterized in order to ease local minima exploration \cite{denil2013predicting, frankle2018lottery}. Thus, various model compression techniques have been proposed for high-performance and/or low-power inference.
For example, pruning techniques remove redundant weights (to zero) without compromising accuracy \cite{optimalbrain}, in order to achieve memory and computation reduction on devices \cite{SHan_2015, sparseVD, suyog_prune, DeepTwist}.
As another model compression technique, non-zero weights can be quantized to fewer bits with comparable model accuracy of full-precision parameters, as discussed in \cite{binaryconnect, rastegariECCV16, Hubara2016, xu2018alternating}.

To achieve even higher compression ratios, pruning and quantization can be combined to form Sparse Quantized Neural Networks (SQNNs).
Intuitively, quantization can leverage parameter pruning since pruning reduces the number of weights to be quantized and quantization loss decreases accordingly \cite{quant_lee}.
Deep compression \cite{deepcompression}, ternary weight networks (TWN) \cite{TWN}, trained ternary quantization (TTQ)
\cite{ternary2017}, and viterbi-based compression \cite{viterbi_quantized, lee2018viterbibased} represent recent efforts to synergistically combine pruning and quantization.

To benefit from sparsity, it is important to (1) represent pruned models in a format with small memory footprint and (2) implement fast computations with sparse matrices as input operands.
Even if reduced SQNNs can be generated with a high pruning rate, it is challenging to gain performance enhancement without an inherently parallel sparse-matrix decoding process during inference.
To illustrate such a challenge, Figure~\ref{intro:cuda} presents DRAM bandwidth, the number of transactions, and execution time of a matrix multiplication using one random $(2048{\times}2048)$ sparse matrix (following CSR format) and a random $(2048{\times}64)$ dense matrix on NVIDIA Tesla V100 (supported by CUDA 9.1).
Because of unbalanced workloads (note that pruning each weight is a somewhat independent and random operation) and additional data for index, sparse matrix computations using CSR format do not offer performance gain as much as sparsity.
Moreover, if pruning rate is not high enough, sparse matrix operations can be even slower than dense matrix operations.

As such, structured and blocked-based pruning techniques \cite{li2016pruning, anwar2017structured, scalpel2017, he2017channel, ye2018rethinking} for DNNs have been proposed to accelerate decoding of sparse matrices using reduced indexing space, as Figure~\ref{intro:pi} shows.
However, coarse-grained pruning associated with reduced indexing space exhibits relatively lower pruning rates compared to unstructured pruning \cite{mao2017exploring}, which masks weights with fine-grained granularity.
In conventional sparse matrix formats representing unstructured pruning (i.e., random weights can be pruned), decoding time can vastly differ if decoding processes are conducted in different blocks simultaneously, as shown in the conventional approach of Figure \ref{intro:recon}.

To enable inherently parallel computations using sparse matrices, this paper proposes a new sparse format.
Our main objective is to remove all pruned weights such that the resulting compression ratio tracks the pruning rate, while maintaining a regular format.
Interestingly, in VLSI testing, proposals for test-data compression have been developed from similar observations, i.e., there are lots of \textit{don't care} bits (= pruned weights in the case of model compression) and the locations of such \textit{don't care} bits seem to be random \cite{survey_test} (the locations of unstructurally pruned weights also seem to be random).
We adopt XOR-gate network, previously used for test-data compression, to decode the compressed bits in a fixed rate during inference, as shown in Figure \ref{intro:recon}.
XOR-gate network is small enough such that we can embed multiple XOR-gate networks to fully utilize memory bandwidth and decode many sparse blocks concurrently.
Correspondingly, we propose an algorithm to find encrypted and compressed data to be fed into XOR-gate network as inputs.





\begin{table*}[t]
\begin{center}  
\begin{tabular}{c|ccc}
\Xhline{2\arrayrulewidth}
 & CSR Format & Viterbi-based Compr. & Proposed \\
\Xhline{2\arrayrulewidth}
Encryption & No & Yes & Yes \\ \hline
Load Balance & Uneven & Even & Even \\ \hline
Decoding Rate & Variable & Fixed & Fixed \\ \hline
Parallelism Limited by & Uneven Sparsity & Number of Decoders & Number of Decoders \\ \hline
Memory Access Pattern & Irregular (Gather-Scatter) &Regular & Regular \\ \hline
\mr{2}{Compressed \\Memory Bandwidth} & \mr{2}{Depends on\\on-chip Buffer Structure} & \mr{2}{1 bit/decoder} & \mr{2}{Multi-bits/decoder} \\
& & & \\ \hline
\mr{2}{HW Resource \\for a Decoder} & \mr{2}{Large Buffer \\to improve load balance} & \mr{2}{XOR gates \\ and Flip-Flops} & \mr{2}{XOR gates \\ only} \\ 
& & & \\ \hline

\Xhline{2\arrayrulewidth}
\end{tabular}
\end{center}
\caption{Comparisons of CSR, Viterbi, and our proposed representation. For all of these representation schemes, compression ratio is upper-bounded by sparsity.} 
\label{table:comparison}
\end{table*}

\section{Related Works and Comparison}
In this section, we introduce two previous approaches to represent sparse matrices. Table \ref{table:comparison} describes CSR format, Viterbi-based index format, and our proposed method.

\textbf{Compressed Sparse Row (CSR)}: Deep compression \cite{deepcompression} utilizes the Compressed Sparse Row (CSR) format to reduce memory footprint on devices.
Unfortunately, CSR formats (including blocked CSR) present irregular data structures not readily supported by highly parallel computing systems such as CPUs and GPUs \cite{lee2018viterbibased}.
Due to uneven sparsity among rows, computation time of algorithms based on CSR is limited by the least sparse row \cite{zhou2018cambricon}, as illustrated in Figure \ref{intro:recon}. 
Although \cite{EIE} suggested hardware supports via a large buffer to improve load balancing, performance is still determined by the lowest pruning rate of a particular row.
In contrast, our scheme provides a perfectly structured format of weights after compression such that high parallelism is maintained. 

\textbf{Viterbi Approaches}:
Viterbi-based compression \cite{lee2018viterbibased, viterbi_quantized} attempts to compress pruning-index data and quantized weights with a fixed compression ratio using \textit{don't care} bits, similar to our approach.
Quantized weights can be compressed by using the Viterbi algorithm to obtain a sequence of inputs to be fed into Viterbi encoders (one bit per cycle).
Because only one bit is accepted for each Viterbi encoder, only an integer number (=number of Viterbi encoder outputs) is permitted as a compression ratio, while our proposed scheme allows any rational numbers.

Because only one bit is used as inputs for Viterbi encoders, Viterbi-based approaches require large hardware resources.
For example, if a memory allows 1024 bits per cycle of bandwidth, then 1024 Viterbi encoders are required, where each Viterbi encoder entails multiple Flip-Flops to support sequence detection.
On the other hand, our proposed scheme is resource-efficient to support large memory bandwidth because Flip-Flops are unnecessary.

\section{Proposed Weight Representation for Structured Compression} 
\label{sect:comp_quant} 
Test-data compression usually generates random numbers as outputs using the input data as seed values.
The outputs (test data containing \textit{don't care} bits) can be compressed successfully if such outputs can be generated by the random number generator using at least one particular input seed data (which is the compressed test data).
It is well known that memory reduction can be as high as the portion of \textit{don't care} bits \cite{Orailoglu, survey_test} if randomness is good enough.
Test data compression and SQNNs with fine-grained pruning share the following properties: 1) Parameter pruning induces \textit{don't care} values as much as pruning rates and 2) If a weight is unpruned, then each quantization bit is assigned to 0 or 1 with equal probability \cite{viterbi_quantized}.
In this section, we propose a new weight representation method exploiting such shared properties while fitting high memory bandwidth requirements and lossless compression to maintain accuracy.

\subsection{Encryption and Decryption}
\begin{figure*}[t]
\begin{center}
\centerline{\includegraphics[width=0.8\linewidth]{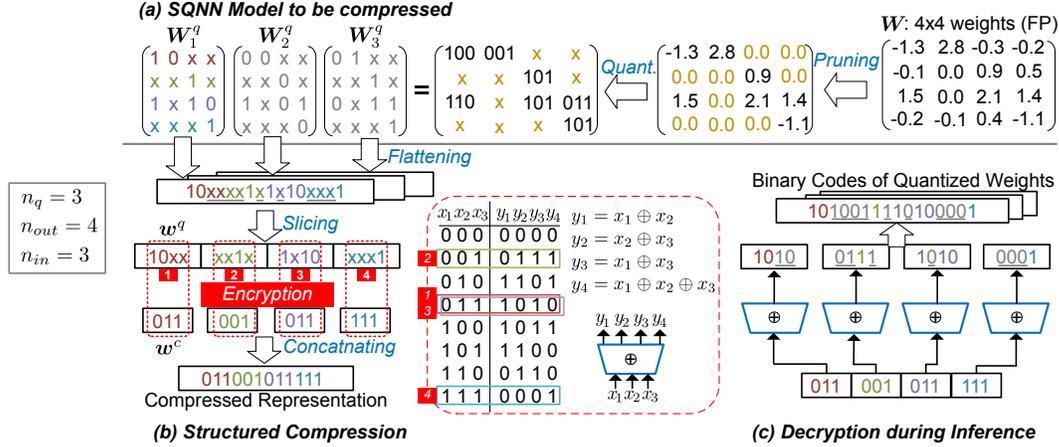}}
\caption{An example illustrating the overall procedures of our proposed method using (4$\times$4) full-precision weights. (a) We assume that a (4$\times$4) weight matrix ($\displaystyle \mW$) is pruned and then quantized into 3 bits. (b) Quantized weights ($\displaystyle \mW^q_{1}, \mW^q_{2}$, and $\mW^q_{3}$) are encrypted by using XOR-gate network which can be formulated as 4 XOR-based equations with 3 inputs ($x_1, x_2$, and $x_3$). We can assign a 3-bit encrypted vector to each sliced 4-bit vector through a look-up table constructed by all possible XOR-gate network input/output pairs. (c) During inference on devices, quantized weights are produced through decryption (that can be best implemented by ASIC or FPGA) from compactly encrypted weights. Note that compared with $\displaystyle \mW^q_{1}, \mW^q_{2}$, and $\mW^q_{3}$, decryption yields new quntized weights in which \textit{care} bits are matched and \textit{don't care} bits are randomly filled.}
\label{quant:bigoverall}
\end{center}
\end{figure*}
We use an XOR-gate network as a random number generator due to its simple design and strong compression capability (such a generator is not desirable for test-data compression because it requires too many input bits).
Suppose that a real-number weight matrix $\mW$ is quantized to be binary matrices $\displaystyle \mW_i^q$ ($1{\le} i{\le} n_q$) with $n_q$ as the number of bits for quantization.
As the first step of our encryption algorithm, we reshape each binary matrix $\displaystyle \mW_i^q$ to be a 1D vector, which is then evenly divided into smaller vector sequences of $n_{out}$ size.
Then, each of the evenly divided vectors, $\vw^q$, including \textit{don't care} bits is encrypted to be a small vector $\vw^c$ (of $n_{in}$ size) without any \textit{don't care} bits.
Through the XOR-gate network, each encrypted vector $\vw^c$ is decrypted to be original bits consisting of correct \textit{care} bits and randomly filled \textit{don't care} bits with respect to $\vw^q$.
Figure \ref{quant:bigoverall} illustrates encryption and decryption procedure examples using a weight matrix.
A 4D tensor (e.g. conv layers) can also be encrypted through the same procedures after flattening.

\begin{figure}
\begin{center}
\centerline{\includegraphics[width=1.0\columnwidth]{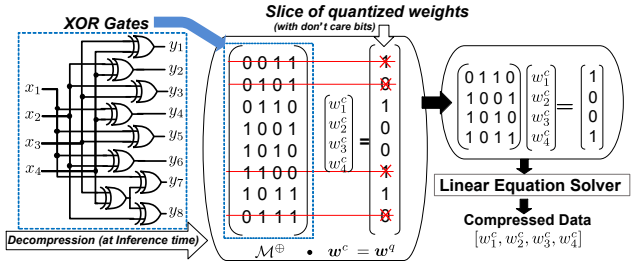}}
\caption{Given a fixed matrix $\displaystyle \mathcal{\mM}$ ($n_{in}{=}4$ and $n_{out}{=}8$) representing the XOR-gate network, encrypting $\displaystyle \vw^q$ is equivalent to solving $\displaystyle \mathcal{\mM}^{\oplus}\vw^c{=}\vw^q$, which can be simplified after removing equations associated with \textit{don't care} bits.}
\label{quant:overall}
\end{center}
\end{figure}

The structure of XOR-gate network is fixed during the entire process and, as depicted in Figure \ref{quant:overall}, can be described as a binary matrix $\displaystyle \mathcal{\mM}^{\oplus}{\in}\{0,1\}^{n_{out}{\times}n_{in}}$ over \textbf{Galois Field} with two elements, $GF(2)$, using the connectivity information between the input vector $\displaystyle \vw^c{\in}\{0,1\}^{n_{in}}$ (compressed and encrypted weights) and $\displaystyle \vw^q{\in}\{0,x,1\}^{n_{out}}$.
Note that $\displaystyle \mathcal{\mM}$ is pre-determined and designed in a way that each element is randomly assigned to 0 or 1 with the same probability.

We intend to generate a random output vector using a seed vector $\displaystyle \vw^c$ while targeting as many \textit{care} bits of $\displaystyle \vw^q$ as possible.
In order to increase the number of successfully matched \textit{care} bits, XOR-gate network should be able to generate various random outputs.
In other words, when the sizes of a seed vector and an output vector are given as $n_{in}$ and $n_{out}$ respectively, all possible $2^{n_{in}}$ outputs need to be well distributed in $2^{n_{out}}$ solution space.

Before discussing how to choose $n_{in}$ and $n_{out}$, let us first study how to find a seed vector $\displaystyle \vw^c$, given $\displaystyle \vw^q$.
As shown in Figure \ref{quant:overall}, the overall operation can be expressed by linear equations $\displaystyle \mathcal{\mM}^{\oplus}\vw^c{=}\vw^q$ over $GF(2)$.
Note that linear equations associated with \textit{don't care} bits in $\displaystyle \vw^q$ can be ignored, because decryption through XOR-gate network can produce any bits in the locations of \textit{don't care} bits.
By deleting unnecessary linear equations, the original equations can be reduced as below:
\begin{fleqn}[\parindent]
\begin{equation}
    \displaystyle \hat{\mathcal{\mathcal{\mM}}}^{\oplus}\vw^c=\vw^q_{\{i_1,...,i_k\}}, \mathord{\mathrm{where}} \\
\end{equation}
\end{fleqn}
\begin{itemize}[noitemsep]
    \item $\{i_1,...,i_k\}$ is a set of indices indicating \textit{care} bits of each vector $\vw^q$ ($0 \le k \le n_{out}, 1\le i_k \le n_{out}$).
    \item $\displaystyle \hat{\mathcal{\mM}}^{\oplus} := \mathcal{\mM}^{\oplus}\left[ i_1,...,i_k; 1,...,n_{in}\right]$
\end{itemize}

For example, since only 4 \textit{care} bits ($\vw^q_{\{3,4,5,7\}}$) exist in Figure \ref{quant:overall}, the (8$\times$4) matrix $\mathcal{\mM}^{\oplus}$ is reduced to a (4$\times$4) matrix, $\hat{\mathcal{\mM}}^{\oplus}$, by removing 1\textsuperscript{st}, 2\textsuperscript{nd}, 6\textsuperscript{th}, and 8\textsuperscript{th} rows.

Given the pruning rate $S$, $\displaystyle \vw^q$ contains $n_{out} {\times} (1{-}S)$ \textit{care} bits on average. 
Assuming that $n_{out} {\times} (1{-}S)$ equations are independent and non-trivial, the required number of seed inputs ($n_{in}$) can be as small as $n_{out} {\times} (1{-}S)$, wherein the compression ratio $n_{out}/n_{in}$ becomes ${1}/(1{-}S)$.
As a result, higher pruning rates lead to higher compression ratios.
However, note that the linear equations may not have a corresponding solution when there are too many `local' \textit{care} bits or there are conflicting equations for a given vector $\displaystyle \vw^q$.

\subsection{Extra Patches for Lossless Compression}

\begin{figure}
\begin{center}
\centerline{\includegraphics[width=0.7\columnwidth]{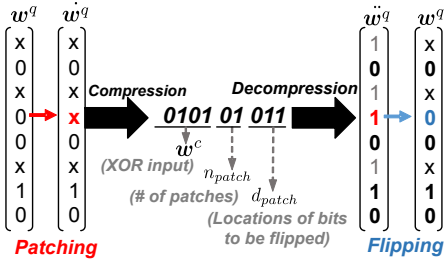}}
\caption{For every $\displaystyle \vw^c$, additional $n_{patch}$ and $d_{patch}$ are attached to match all \textit{care} bits.}
\label{quant:patch}
\end{center}

\end{figure}

In order to keep our proposed SQNNs representation lossless, we add extra bits to correct unavoidable errors, i.e., patching.
An unsolvable linear equation implies that the XOR random number generator cannot produce one or more matched \textit{care} bits of $\displaystyle \vw$.
To resolve such an occasion, we can replace one or more \textit{care} bits of $\displaystyle \vw^q$ with \textit{don't care} bits to remove conflicting linear equations, as depicted in Figure~\ref{quant:patch}.
We record the locations of replacements as $d_{patch}$ which can be used to recover the original \textit{care} bits of $\displaystyle \vw$ by {\it flipping} the corresponding bits during decryption.
For every $\displaystyle \vw^q$, $n_{patch}$ indicates the number of replacements for each $\displaystyle \vw^c$.
Since $n_{patch}$ is always scanned prior to decryption using $\displaystyle \vw^c$, the same number of bits is reserved to represent $n_{patch}$ for all encrypted vectors in order to maintain a regular compressed format.
On the other hand, the size of $d_{patch}$ can be different for each $\vw^q$ (overall parallelism is not disrupted by different $d_{patch}$ sizes as flipping occurs infrequently).

At the expense of $n_{patch}$ and $d_{patch}$, our compression technique can reproduce all \textit{care} bits of $\displaystyle \vw^q$ and, therefore, does not affect accuracy of DNN models and obviates retraining.
In sum, the compressed format includes 1) $\displaystyle \vw^c_{1...l}{\in}\{0,1\}^{n_{in}}$ ($l {=} \lceil\frac{mn}{n_{out}}\rceil$) encrypted from a quantized weight matrix $\displaystyle \mW_i^q {\in}\{0,x,1\}^{m \times n}$ through $\displaystyle \mathcal{\mM}^{\oplus}{\in}\{0,1\}^{n_{out}\times n_{in}}$, 2) $n_{patch}$, and 3) $d_{patch}$.
Hence, the resulting compression ratio $r$ is
\begin{equation}
\label{eq:q1}
\displaystyle r = \frac{{mn}}{({\frac{n_{in}}{n_{out}}}mn + {l \lceil \lg{\max(\vp)}\rceil} + {\sum\limits_{j=1}^{l}{\evp_j}\lceil\lg{n_{out}}\rceil })},
\end{equation}
where $p_j$ is the $j$\textsuperscript{th} $n_{patch}$ and $\displaystyle \vp$ = $\{p_1, p_2, ..., p_l\}$.
Improving $r$ is enabled by increasing $n_{out}/n_{in}$ and decreasing the amount of patches.
We introduce a heuristic patch-searching algorithm to reduce the number of patches while also optimizing $n_{in}$ and $n_{out}$. 
\renewcommand\algorithmiccomment[1]{%
  \hfill \eqparbox{COMMENT}{// \textit{#1}}%
}
\subsection{Experiments Using Synthetic Data}
\begin{algorithm}
   \SetAlgoLined
   \SetKwInOut{Input}{input}
   \SetKwInOut{Output}{output}

   \caption{Patch-Searching Algorithm}
   \label{alg:heu}
   \Input{a set of indices of care bits $\{i_1,...,i_k\}$, \\
          a sliced weight vector $\displaystyle \vw^q \in\{0,x,1\}^{n_{out}}$, \\
          a fixed matrix $\displaystyle \mathcal{\mM}^{\oplus} \in \{0,1\}^{n_{out} \times n_{in}}$}
\begin{algorithmic}[1]
   
   \STATE $mat$ $\leftarrow$ empty matrix which column size is $n_{in}+1$
   
   \FOR[for only not-pruned bits]{$i$ in $\{i_1,...,i_k\}$}
        \STATE Append a row ($\displaystyle \mathcal{\emM}^{\oplus}_{i,1}, ...., \mathcal{\emM}^{\oplus}_{i,n_{in}}, \evw^q_i$) to $mat$
        \STATE $rref$ $\leftarrow$ make\_rref($mat$)
        \IF{$rref$.is\_solved() is False}
            \STATE Remove the last row of $mat$ from $mat$
        \ENDIF
    \ENDFOR
   \STATE Solve linear equations using $mat$ to find $\displaystyle \vw^c$
   \STATE $\displaystyle \ddot{\vw}^q$ $\leftarrow$ $\displaystyle \mathcal{\mM}^{\oplus} \vw^c$
   \STATE Compare $\displaystyle \ddot{\vw}^q$ with $\displaystyle \vw^q$ to produce $n_{patch}$ and $d_{patch}$
   \STATE \textbf{Return} $n_{patch}$, $d_{patch}$, $\displaystyle \vw^c$
\end{algorithmic}
\end{algorithm}

\begin{figure}[t]
\begin{center}
	    \centering
		\includegraphics[width=0.9\columnwidth]{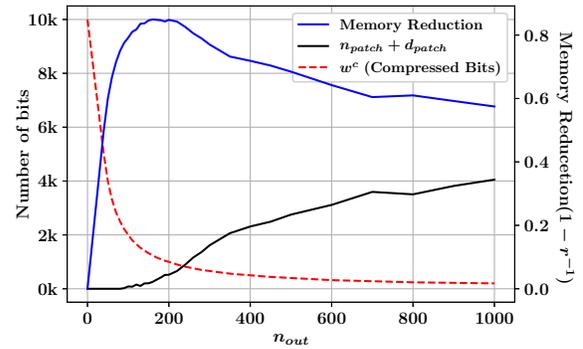}
		\caption{Memory reduction by applying Algorithm \ref{alg:heu} to a random weight matrix with 10,000 elements with various $n_{out}$ (pruning rate $S$=0.9, $n_{in}$=20)}
		\label{quant:grace}
\end{center}

\end{figure}

\begin{figure*}[t]
\begin{center}
    \begin{subfigure}{.32\textwidth}
	    \includegraphics[width=1.0\columnwidth]{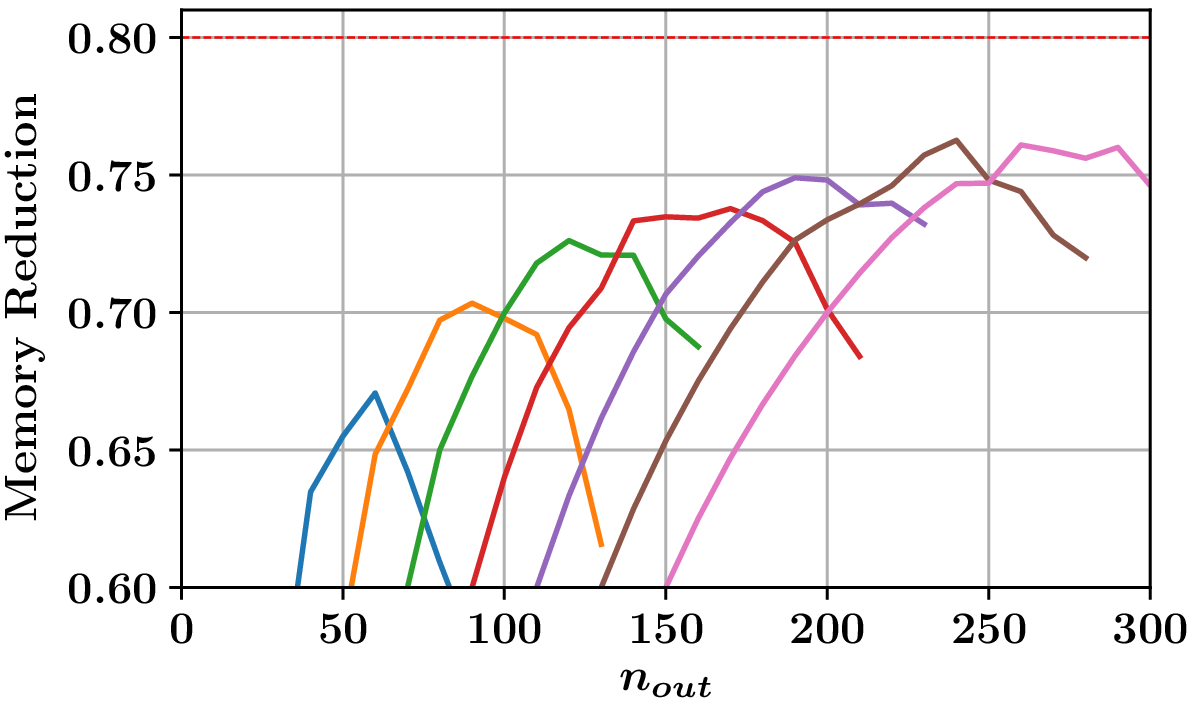}
	    \caption{$S=0.8$}
	\end{subfigure}
	\begin{subfigure}{.32\textwidth}
	    \includegraphics[width=1.0\columnwidth]{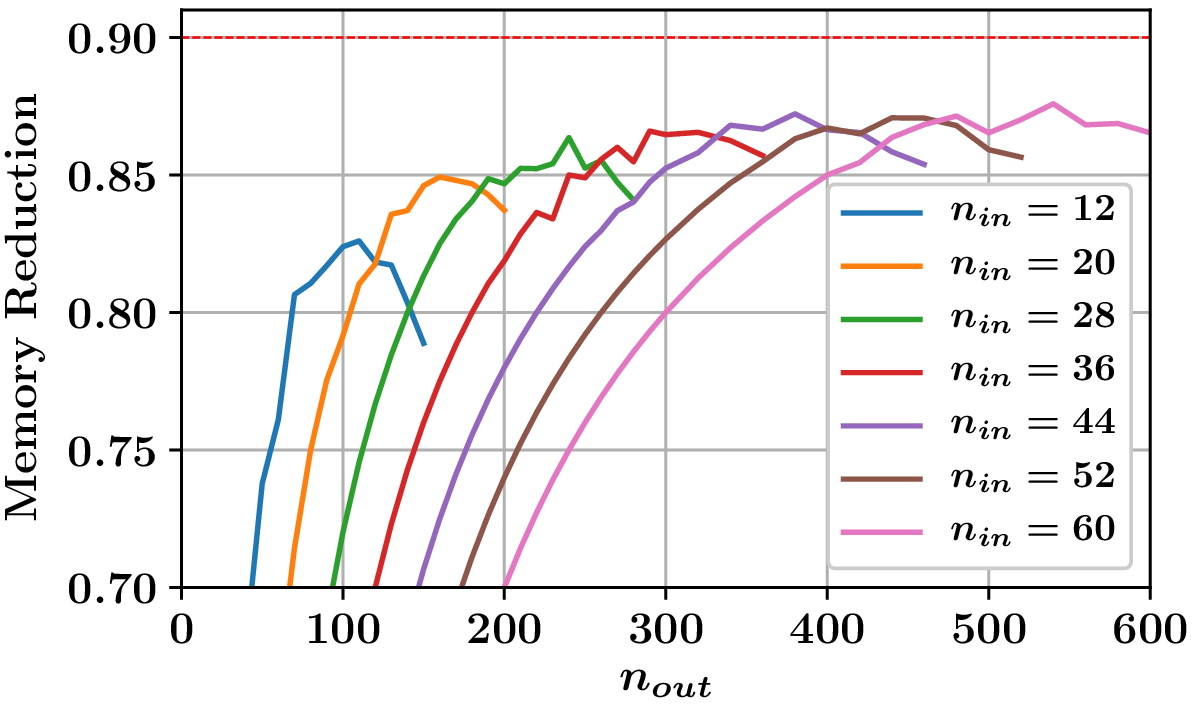}
	    \caption{$S=0.9$}
	\end{subfigure}
	\begin{subfigure}{.32\textwidth}
	    \includegraphics[width=1.0\columnwidth]{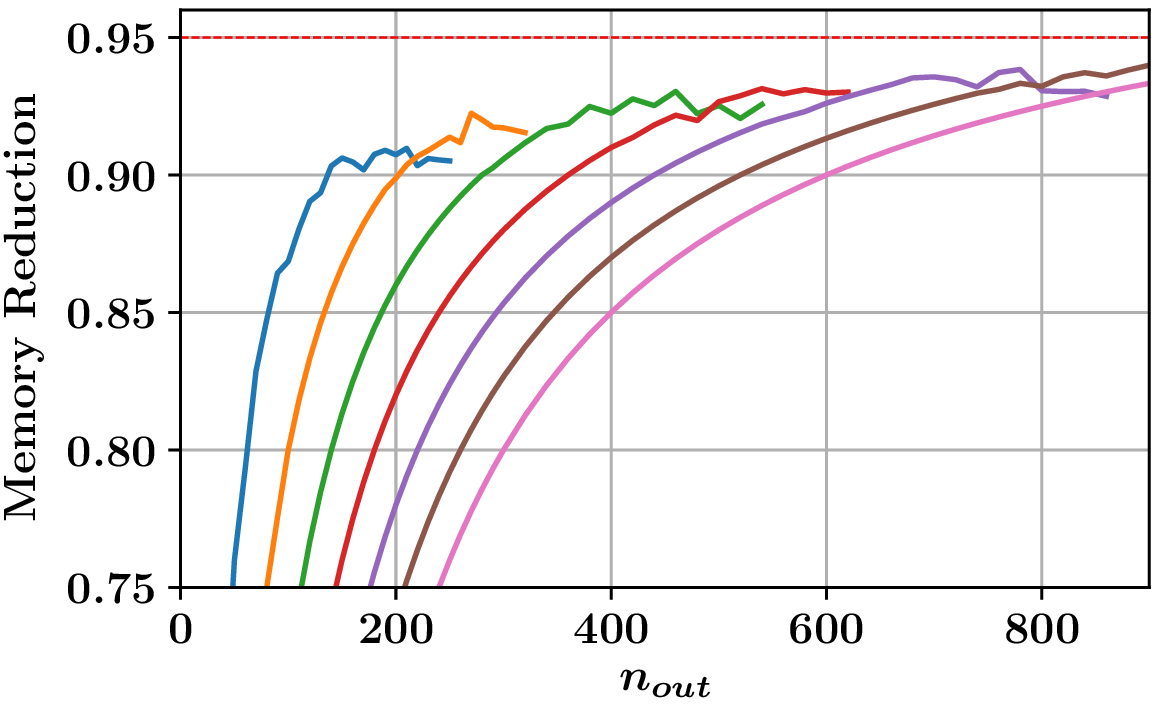}
	    \caption{$S=0.95$}
	\end{subfigure}
	\caption{Memory reduction using various $n_{in}$ and pruning rates. $n_{in}$ ranges from 12 to 60. Each line is stopped when the memory reduction begins to fall.}
	\label{quant:result1}
\end{center}

\end{figure*}

\begin{figure}[t]
\begin{center}
	    \includegraphics[width=0.8\columnwidth]{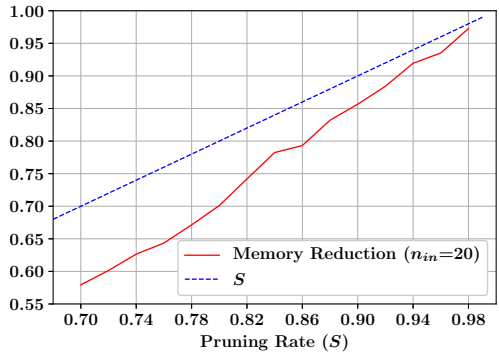}
		\caption{Graphs of memory reduction using $n_{in}$=20 (red line) and $S$ (blue line). The gap between those two graphs is reduced with higher pruning rate.}
		\label{quant:result2}
\end{center}

\end{figure}

An exhaustive search of patches requires exponential-time complexity even though such a method minimizes the number of patches.
Algorithm \ref{alg:heu} is a heuristic algorithm to search encrypted bits including $n_{patch}$ and $d_{patch}$ for $\vw^q$ in $\mW$.
The algorithm incrementally enlarges the reduced linear equation by including a \textit{care} bit only when the enlarged equation is still solvable.
Note that make\_rref() in Algorithm \ref{alg:heu} generates a reduced row-echelon form to quickly verify that the linear equations are solvable.
If adding a certain \textit{care} bit makes the equations unsolvable, then a \textit{don't care} bit takes its place, and $n_{patch}$ and $d_{patch}$ are updated accordingly. 
Although Algorithm \ref{alg:heu} yields more replacement of \textit{care} bits than an exhaustive search (by up to 10\% from our extensive experiments), our simple algorithm has time complexity of $O(n_{out})$, which is much faster.

To investigate the compression capability of our proposed scheme, we evaluate a large random weight matrix of 10,000 elements where each element becomes a \textit{don't care} bit with the probability of 0.9 (=sparsity or pruning rate).
If an element is a \textit{care} bit, then a 0 or 1 is assigned with the same probability.
Notice that randomness of locations in \textit{don't care} bits is a feature of fine-grained pruning methods and assignment of 0 or 1 to weights with the same probability is obtainable using well-balanced quantization techniques \cite{viterbi_quantized, lee2018viterbibased}.

When $n_{in}$ is fixed, the optimal $n_{out}$ maximizes the memory reduction offered by Algorithm \ref{alg:heu}.
Figure \ref{quant:grace} plots the corresponding memory reduction (=$1{-}r^{-1}$) from $\displaystyle \mW^q$ on the right axis and the amount of $\displaystyle \vw^c$ and $n_{patch}$ {+} $d_{patch}$ on the left axis across a range of $n_{out}$ values when $n_{in}{=}20$.
From Figure \ref{quant:grace}, it is clear that there exists a trade-off between the size of $\vw^c$ and the sizes of $n_{patch}$ and $d_{patch}$.
Increasing $n_{out}$ rapidly reduces $\displaystyle \vw^c$ while $n_{patch}$ and $d_{patch}$ grow gradually.
The highest memory reduction ($\approx 0.83$) is achieved when $n_{out}$ is almost 200, which agrees with the observation that maximum compression is constrained by the relative number of \textit{care} bits \cite{survey_test}.
Consequently, the resulting compression ratio approaches $1/(1{-}S)$.

Given the relationship above, we can now optimize $n_{in}$.
Figure \ref{quant:result1} compares memory reduction for various $n_{out}$ across different values of $n_{in}$.
The resulting trend suggests that higher $n_{in}$ yields more memory reduction.
This is because increasing the number of bits used as seed values for the XOR-gate random number generator enables a larger solution space and, as a result, fewer $d_{patch}$ are needed as $n_{in}$ increases. 
The large solution space is especially useful when \textit{don't care} bits are not evenly distributed throughout $\displaystyle \mW^q$.
Lastly, $n_{in}$ is constrained by the maximum computation time available to run Algorithm \ref{alg:heu}.

Figure \ref{quant:result2} presents the relationship between pruning rate $S$ and memory reduction when $n_{in}{=}20$ and sweeping $S$.
Since high pruning rates translate to fewer \textit{care} bits and relatively fewer $d_{patch}$, Figure \ref{quant:result2} confirms that memory reduction reaches $S$ as $S$ increases.
In other words, maximizing pruning rate is key to compressing quantized weights with high compression ratio.
In comparison, ternary quantization usually induces a lower pruning rate \cite{ternary2017, TWN}.
Our proposed representation is best implemented by pruning first to maximize pruning rate and then quantizing the weights.

\section{Experiments on various SQNNs}
\begin{table*}[t]
\begin{center}
\begin{tabular}{ccc|c|ccc}
\Xhline{2\arrayrulewidth}
  \multicolumn{3}{c|}{Target Model} & Pre-trained & \multicolumn{3}{c}{Pruning and Quantization} \\ 
  \hline
  \mr{1}{Model} & \mr{1}{DataSet} & \mr{1}{Size} & \mr{1}{Acc.} & \mr{1}{$S$} & \mr{1}{$n_q$-bit} & \mr{1}{Acc.} \\
\Xhline{2\arrayrulewidth}
       \mr{1}{LeNet5 (FC1)} & \mr{1}{MNIST} & \mr{1}{800$\times$500} & \mr{1}{99.1\%} & \mr{1}{0.95} & \mr{1}{1-bit} & \mr{1}{99.1\%} \\
  \hline
      \mr{1}{AlexNet (FC5, FC6)} & \mr{1}{ImageNet} & \mr{1}{9K$\times$4K, 4K$\times$4K} & \mr{1}{57.6\% (T1)}& \mr{1}{0.91} & \mr{1}{1-bit} & \mr{1}{55.9\% (T1)} \\
  \hline
       \mr{1}{ResNet32 (Conv Layers)} & \mr{1}{CIFAR10} & \mr{1}{460.76K} & \mr{1}{92.5\%} & \mr{1}{0.70} & \mr{1}{2-bit} & \mr{1}{91.6\%} \\
 \hline
    \mr{1}{LSTM} & \mr{1}{PTB} & \mr{1}{6.41M} & \mr{1}{89.6 PPW} & \mr{1}{0.60} & \mr{1}{2-bit} & \mr{1}{93.9 PPW} \\
\Xhline{2\arrayrulewidth}
\end{tabular}
\end{center}

\caption{Descriptions of models to be compressed by our proposed method. The model accuracy after parameter pruning and quantization is obtained by a binary-index factorization \cite{ourBMF} and alternating multi-bit quantization
\label{table:results}
\cite{xu2018alternating}.}

\end{table*}

\begin{figure*}[t]
\begin{center}
\centerline{\includegraphics[width=0.8\textwidth]{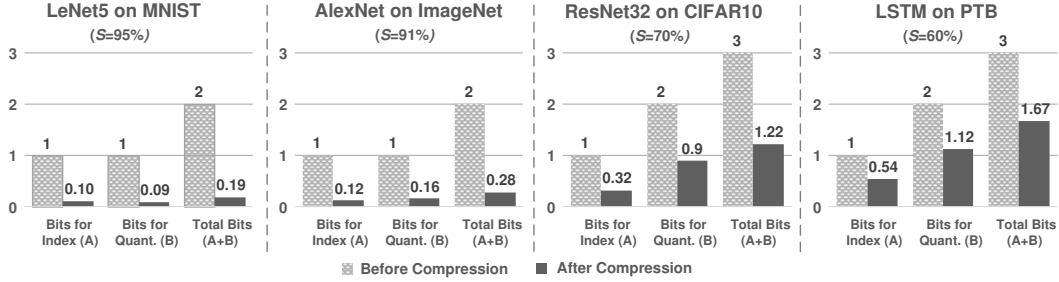}}
\caption{The number of bits to represent each weight for the models in Table \ref{table:results} using our proposed SQNNs format. (A) means the number of bits for index (compressed by binary-index matrix factorization introduced in \cite{ourBMF}). (B) indicates the number of bits for quantization by our proposed compression technique. Overall, we gain additional 2-11$\times$ memory footprint reduction according to sparsity. Note that memory overhead due to XOR-gate network is negligible because a relatively small XOR-gate network is pre-determined and fixed in advance.} 
\label{graph}
\end{center}
\end{figure*}
In this section, we show experimental results of the proposed representation for four popular datasets: MNIST, ImageNet \cite{imagenet}, CIFAR10 \cite{cifar10}, and Penn Tree Bank \cite{ptb}.
Though the compression ratio ideally reaches $1/(1{-}S)$, the actual results may not, because \textit{don't care} bits can be less evenly distributed than the synthetic data we used for Section 3.3.
Hence, we suggest several additional techniques in this section to handle uneven distributions.

Weights are pruned by the mask layer generated by binary-index matrix factorization \cite{ourBMF} after pre-training, and then retrained.
Quantization is performed by following the technique proposed in \cite{DeepTwist} and \cite{efficient_quant}, where quantization-aware optimization is performed based on the quantization method from \cite{xu2018alternating}.
The number of bits per weight required by our method is compared with the case of $n_q$-bit quantization with an additional 1-bit indicating pruning index (e.g., ternary quantization consists of 1-bit quantization and 1-bit pruning indication per weight).


We first tested our representation using the LeNet-5 model on MNIST. 
LeNet-5 consists of two convolutional layers and two fully-connected layers \cite{SHan_2015, DeepTwist}.
Since the FC1 layer dominates the memory footprint (93\%), we focus only on the FC1 layer whose parameters can be pruned by 95\%.
With our proposed method, the FC1 layer is effectively represented by only 0.19 bits per weight, which is $11\times$ smaller than ternary quantization, as Figure \ref{graph} shows.
We also tested our proposed compression techniques on large-scale models and datasets, namely, AlexNet \cite{alexnet} on the ImageNet dataset.
We focused on compressing FC5 and FC6 fully-connected layers occupying $\sim$90\% of the total model size for AlexNet.
Both layers are pruned by a pruning rate of 91\% \cite{SHan_2015} using binary-index matrix factorization \cite{ourBMF} and compressed by 1-bit quantization.
The high pruning rate lets us compress the quantized weights by ${\sim}7\times$.
Overall, FC5 and FC6 layers for AlexNet require only 0.28 bits per weight, which is substantially less than 2 bits per weight required by ternary quantization.

We further verify our compression techniques using ResNet32 \cite{resnet} on the CIFAR10 dataset with a baseline accuracy of 92.5\%. 
The model is pruned and quantized to 2 bits, reaching 91.6\% accuracy after retraining. Further compression with our proposed SQNN format yields 1.22 bits per weight, while 3 bits would be required without our proposed compression techniques.

Additionally, an RNN model with one LSTM layer of size 300 \cite{xu2018alternating} on the PTB dataset, with performance measured by using Perplexity Per Word, is compressed by our representation.
Following our proposed representation scheme along with pruning and 2-bit quantization, such PTB LSTM model requires only 1.67 bits per weight.

For various types of layers, our proposed technique, supported by weight sparsity, provides additional compression over ternary quantization. 
Compression ratios can be further improved by using more advanced pruning and quantization methods (e.g., \cite{HitNet, DNS}) since the principles of our compression methods do not rely on the specific pruning and quantization methods used.

\section{Discussion}

\subsection{Variation on Execution Time}

\begin{figure}[t]
\begin{center}
	    \includegraphics[width=0.8\columnwidth]{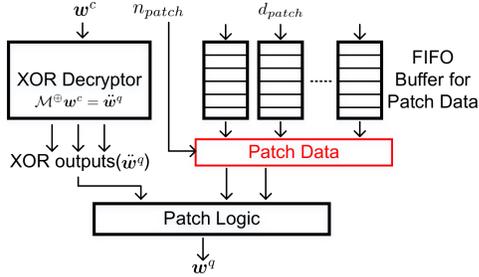}
		\caption{Patch data process structure corresponding to each XOR-gate network with a multi-bank FIFO to store patch data.}
		\label{fig:FIFO_architecture}
\end{center}

\end{figure}

While decryption process through XOR-gate network provides a fixed output rate (thus, high parallelism), if all mismatched bits are supposed to be corrected by patches, then entire decoding (including decryption and patch correction) may result in variation in execution time due to irregular patch size.
In order to mitigate variation in patch process, we first assume that patch data $d_{patch}$ structure is decoupled from encrypted weights such that $d_{patch}$ is given as a stream data to be stored into buffers in a fixed rate.
Then, for each XOR decryption cycle, $d_{patch}$ as much as $n_{patch}$ is read from buffers and used to fix XOR outputs.
In Figure~\ref{fig:FIFO_architecture}, $d_{patch}$ is stored into or loaded from FIFO buffers when the number of FIFO banks is $n_{\mathit{F\!I\!F\!O}}$.
If $d_{patch}$ is needed, FIFO buffers deliver $d_{patch}$ to patch process logic while available $d_{patch}$ throughput for load/store is determined by the number of FIFO banks.
Decoding process can be stalled when the FIFO is either full or empty if temporal $d_{patch}$ consumption rate is too low or too high.

\begin{figure}[t]
\begin{center}
	    \includegraphics[width=1.0\columnwidth]{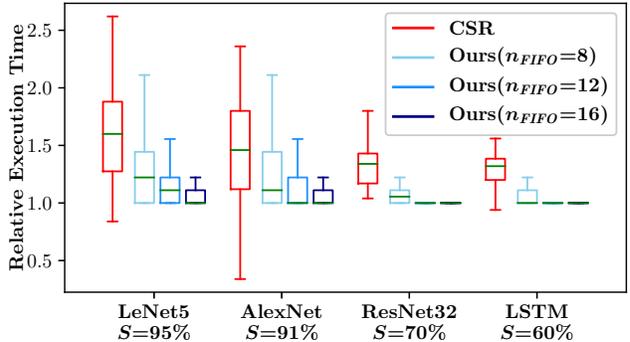}
		\caption{Relative execution time with CSR and the proposed scheme with various $n_{\mathit{F\!I\!F\!O}}$ size. 1.0 in y-axis means no pruning rate variation in each row in the case of CSR format, or no stall due to limited $d_{patch}$ load bandwidth in the case of the proposed scheme.} 
		\label{fig:Patch_Execution_time}
\end{center}

\end{figure}

Figure~\ref{fig:Patch_Execution_time} presents relative execution time using CSR format or the proposed scheme with different $n_{\mathit{F\!I\!F\!O}}$ configurations.
FIFO size can be small enough (say, 256 entries) to tolerate temporal high peak $d_{patch}$ usage.
Hence, in Figure~\ref{fig:Patch_Execution_time}, stalls in the proposed scheme are due to high $d_{patch}$ throughput demands along with large $n_{patch}$.
Note that CSR format yields high variations in execution time because each row exhibits various numbers of unpruned weights.
On the other hand, in the case of the proposed scheme, increasing $n_{\mathit{F\!I\!F\!O}}$ (at the cost of additional hardware design) enhances $d_{patch}$ throughput, and thus, reduces the number of stalls incurred by limited $d_{patch}$ bandwidth of FIFOs.
In sum, reasonable $n_{\mathit{F\!I\!F\!O}}$ size significantly reduces execution time variations, which has been a major bottleneck in implementing fine-grained pruning schemes, as demonstrated in Figure \ref{intro:cuda}.

\subsection{Practical Techniques to Reduce \boldmath$n_{patch}$}
If $n_{out}$ is large enough, patching overhead is not supposed to disrupt the parallel decoding, ideally. 
However, even for large $n_{out}$, if the nonuniformity of pruning rates is observed over a wide range within a matrix, $n_{patch}$ may considerably increase. 
Large $n_{patch}$, then, leads to not only degraded compression ratio compared with synthetic data experiments, but also deteriorated parallelism in the decoding process.
The following techniques can be considered to reduce $n_{patch}$. 

\textbf{Blocked \boldmath$n_{patch}$ Assignment}: The compression ratio $r$ in the Eq.~(8) of Section 3.2 is affected by the maximum of $\{p_1, p_2, ..., p_l\}$. 
Note that a particular vector $\displaystyle \vw$ may have an exceptionally large number of \textit{care} bits.
In such a case, even if a quantized matrix $\displaystyle \mW^q$ consists of mostly \textit{don't care} bits and few patches are needed, all of the compressed vectors $\displaystyle \vw^c$ must employ a large number of bits to track the number of patches. 
To mitigate such a problem and enhance the compression ratio $r$, we divide a binary matrix $\displaystyle \mW^q$ into several blocks, and then $\displaystyle \max(\vp)$ is obtained in each block independently. 
Different $n_{patch}$ is assigned to each block to reduce the overall $n_{patch}$ data size.

\textbf{Minimizing \boldmath$n_{patch}$ for Small \boldmath$n_{in}$}:  
One simple patch-minimizing algorithm is to list all possible $2^{n_{in}}$ inputs (for $\vw^c$) and corresponding outputs through $\mathcal{\mM}^{\oplus}$ on memory and find a particular $\vw^c$ that minimizes the number of patches.
At the cost of high space complexity and memory consumption, such an exhaustive optimization guarantees minimized $n_{patch}$.
$n_{in}$ below 30 is a practical value.



\section{Conclusion}
This paper proposes a compressed representation for Sparse Quantized Neural Networks based on an idea used for test-data compression.
Through XOR-gate network and solving linear equations, we can remove most \textit{don't care} bits and a quantized model is further compressed by sparsity.
Since our representation provides a regular compressed-weight format with fixed and high compression ratios, SQNNs enable not only memory footprint reduction but also inference performance improvement due to inherently parallelizable computations.

{\small
\bibliographystyle{ieee_fullname}
\bibliography{egbib}

\begin{thebibliography}{10}\itemsep=-1pt

\bibitem{viterbi_quantized}
Daehyun Ahn, Dongsoo Lee, Taesu Kim, and Jae-Joon Kim.
\newblock Double {Viterbi}: Weight encoding for high compression ratio and fast
  on-chip reconstruction for deep neural network.
\newblock In {\em International Conference on Learning Representations (ICLR)},
  2019.

\bibitem{anwar2017structured}
Sajid Anwar, Kyuyeon Hwang, and Wonyong Sung.
\newblock Structured pruning of deep convolutional neural networks.
\newblock {\em ACM Journal on Emerging Technologies in Computing Systems
  (JETC)}, 13(3):32, 2017.

\bibitem{Orailoglu}
Ismet Bayraktaroglu and Alex Orailoglu.
\newblock Test volume and application time reduction through scan chain
  concealment.
\newblock In {\em Proceedings of the 38th Annual Design Automation Conference},
  2001.

\bibitem{binaryconnect}
Matthieu Courbariaux, Yoshua Bengio, and Jean-Pierre David.
\newblock {BinaryConnect}: Training deep neural networks with binary weights
  during propagations.
\newblock In {\em Advances in Neural Information Processing Systems}, pages
  3123--3131, 2015.

\bibitem{denil2013predicting}
Misha Denil, Babak Shakibi, Laurent Dinh, Nando De~Freitas, et~al.
\newblock Predicting parameters in deep learning.
\newblock In {\em Advances in neural information processing systems}, pages
  2148--2156, 2013.

\bibitem{frankle2018lottery}
Jonathan Frankle and Michael Carbin.
\newblock The lottery ticket hypothesis: Finding sparse, trainable neural
  networks.
\newblock {\em arXiv preprint arXiv:1803.03635}, 2018.

\bibitem{deeplearningbook}
Ian Goodfellow, Yoshua Bengio, and Aaron Courville.
\newblock {\em Deep Learning}.
\newblock MIT Press, 2016.
\newblock \url{http://www.deeplearningbook.org}.

\bibitem{DNS}
Yiwen Guo, Anbang Yao, and Yurong Chen.
\newblock Dynamic network surgery for efficient {DNNs}.
\newblock In {\em Advances in Neural Information Processing Systems}, 2016.

\bibitem{EIE}
Song Han, Xingyu Liu, Huizi Mao, Jing Pu, Ardavan Pedram, Mark~A. Horowitz, and
  William~J. Dally.
\newblock {EIE:} efficient inference engine on compressed deep neural network.
\newblock In {\em Proceedings of the 43rd International Symposium on Computer
  Architecture}, pages 243--254, 2016.

\bibitem{deepcompression}
Song Han, Huizi Mao, and William~J. Dally.
\newblock Deep compression: Compressing deep neural networks with pruning,
  trained quantization and {Huffman} coding.
\newblock In {\em International Conference on Learning Representations (ICLR)},
  2016.

\bibitem{SHan_2015}
Song Han, Jeff Pool, John Tran, and William~J. Dally.
\newblock Learning both weights and connections for efficient neural networks.
\newblock In {\em Advances in Neural Information Processing Systems}, pages
  1135--1143, 2015.

\bibitem{resnet}
Kaiming He, Xiangyu Zhang, Shaoqing Ren, and Jian Sun.
\newblock Deep residual learning for image recognition.
\newblock {\em 2016 IEEE Conference on Computer Vision and Pattern Recognition
  (CVPR)}, pages 770--778, 2016.

\bibitem{he2017channel}
Yihui He, Xiangyu Zhang, and Jian Sun.
\newblock Channel pruning for accelerating very deep neural networks.
\newblock In {\em Proceedings of the IEEE International Conference on Computer
  Vision}, pages 1389--1397, 2017.

\bibitem{Hubara2016}
Itay Hubara, Matthieu Courbariaux, Daniel Soudry, Ran El-Yaniv, and Yoshua
  Bengio.
\newblock Quantized neural networks: training neural networks with low
  precision weights and activations.
\newblock {\em arXiv:1609.07061}, 2016.

\bibitem{efficient_quant}
Parichay Kapoor, Dongsoo Lee, Byeongwook Kim, and Saehyung Lee.
\newblock Computation-efficient quantization method for deep neural networks,
  2019.

\bibitem{cifar10}
Alex Krizhevsky.
\newblock Learning multiple layers of features from tiny images.
\newblock Technical report, 2009.

\bibitem{alexnet}
Alex Krizhevsky, Ilya Sutskever, and Geoffrey~E Hinton.
\newblock Imagenet classification with deep convolutional neural networks.
\newblock In F. Pereira, C.~J.~C. Burges, L. Bottou, and K.~Q. Weinberger,
  editors, {\em Advances in Neural Information Processing Systems 25}, pages
  1097--1105. Curran Associates, Inc., 2012.

\bibitem{optimalbrain}
Yann LeCun, John~S. Denker, and Sara~A. Solla.
\newblock Optimal brain damage.
\newblock In {\em Advances in Neural Information Processing Systems}, pages
  598--605, 1990.

\bibitem{lee2018viterbibased}
Dongsoo Lee, Daehyun Ahn, Taesu Kim, Pierce~I. Chuang, and Jae-Joon Kim.
\newblock Viterbi-based pruning for sparse matrix with fixed and high index
  compression ratio.
\newblock In {\em International Conference on Learning Representations (ICLR)},
  2018.

\bibitem{DeepTwist}
Dongsoo Lee, Parichay Kapoor, and Byeongwook Kim.
\newblock Deeptwist: Learning model compression via occasional weight
  distortion.
\newblock {\em arXiv:1810.12823}, 2018.

\bibitem{quant_lee}
Dongsoo Lee and Byeongwook Kim.
\newblock Retraining-based iterative weight quantization for deep neural
  networks.
\newblock {\em arXiv:1805.11233}, 2018.

\bibitem{ourBMF}
Dongsoo Lee, Se~Jung Kwon, Parichay Kapoor, Byeongwook Kim, and Gu-Yeon Wei.
\newblock Network pruning for low-rank binary indexing.
\newblock {\em arXiv:1905.05686}, 2019.

\bibitem{TWN}
Fengfu Li and Bin Liu.
\newblock Ternary weight networks.
\newblock {\em arXiv:1605.04711}, 2016.

\bibitem{li2016pruning}
Hao Li, Asim Kadav, Igor Durdanovic, Hanan Samet, and Hans~Peter Graf.
\newblock Pruning filters for efficient convnets.
\newblock In {\em International Conference on Learning Representations}, 2017.

\bibitem{mao2017exploring}
Huizi Mao, Song Han, Jeff Pool, Wenshuo Li, Xingyu Liu, Yu Wang, and William~J
  Dally.
\newblock Exploring the regularity of sparse structure in convolutional neural
  networks.
\newblock {\em arXiv preprint arXiv:1705.08922}, 2017.

\bibitem{ptb}
Mitchell Marcus, Grace Kim, Mary~Ann Marcinkiewicz, Robert MacIntyre, Ann Bies,
  Mark Ferguson, Karen Katz, and Britta Schasberger.
\newblock The penn treebank: Annotating predicate argument structure.
\newblock In {\em Proceedings of the Workshop on Human Language Technology},
  HLT '94, pages 114--119, Stroudsburg, PA, USA, 1994. Association for
  Computational Linguistics.

\bibitem{sparseVD}
Dmitry Molchanov, Arsenii Ashukha, and Dmitry~P. Vetrov.
\newblock Variational dropout sparsifies deep neural networks.
\newblock In {\em International Conference on Machine Learning ({ICML})}, pages
  2498--2507, 2017.

\bibitem{rastegariECCV16}
Mohammad Rastegari, Vicente Ordonez, Joseph Redmon, and Ali Farhadi.
\newblock {XNOR-Net}: Imagenet classification using binary convolutional neural
  networks.
\newblock In {\em ECCV}, 2016.

\bibitem{imagenet}
Olga Russakovsky, Jia Deng, Hao Su, Jonathan Krause, Sanjeev Satheesh, Sean Ma,
  Zhiheng Huang, Andrej Karpathy, Aditya Khosla, Michael Bernstein,
  Alexander~C. Berg, and Li Fei-Fei.
\newblock {ImageNet Large Scale Visual Recognition Challenge}.
\newblock {\em International Journal of Computer Vision (IJCV)},
  115(3):211--252, 2015.

\bibitem{survey_test}
Nur~A. Touba.
\newblock Survey of test vector compression techniques.
\newblock {\em IEEE Design \& Test of Computers}, 23:294--303, 2006.

\bibitem{HitNet}
Peiqi Wang, Xinfeng Xie, Lei Deng, Guoqi Li, Dongsheng Wang, and Yuan Xie.
\newblock {HitNet}: Hybrid ternary recurrent neural network.
\newblock In {\em Advances in Neural Information Processing Systems}, 2018.

\bibitem{xu2018alternating}
Chen Xu, Jianqiang Yao, Zhouchen Lin, Wenwu Ou, Yuanbin Cao, Zhirong Wang, and
  Hongbin Zha.
\newblock Alternating multi-bit quantization for recurrent neural networks.
\newblock In {\em International Conference on Learning Representations (ICLR)},
  2018.

\bibitem{ye2018rethinking}
Jianbo Ye, Xin Lu, Zhe Lin, and James~Z Wang.
\newblock Rethinking the smaller-norm-less-informative assumption in channel
  pruning of convolution layers.
\newblock {\em arXiv preprint arXiv:1802.00124}, 2018.

\bibitem{scalpel2017}
Jiecao Yu, Andrew Lukefahr, David Palframan, Ganesh Dasika, Reetuparna Das, and
  Scott Mahlke.
\newblock Scalpel: Customizing {DNN} pruning to the underlying hardware
  parallelism.
\newblock In {\em Proceedings of the 44th Annual International Symposium on
  Computer Architecture}, pages 548--560, 2017.

\bibitem{zhou2018cambricon}
Xuda Zhou, Zidong Du, Qi Guo, Shaoli Liu, Chengsi Liu, Chao Wang, Xuehai Zhou,
  Ling Li, Tianshi Chen, and Yunji Chen.
\newblock Cambricon-s: Addressing irregularity in sparse neural networks
  through a cooperative software/hardware approach.
\newblock In {\em 2018 51st Annual IEEE/ACM International Symposium on
  Microarchitecture (MICRO)}, pages 15--28. IEEE, 2018.

\bibitem{ternary2017}
Chenzhuo Zhu, Song Han, Huizi Mao, and William~J. Dally.
\newblock Trained ternary quantization.
\newblock In {\em International Conference on Learning Representations (ICLR)},
  2017.

\bibitem{suyog_prune}
Michael Zhu and Suyog Gupta.
\newblock To prune, or not to prune: exploring the efficacy of pruning for
  model compression.
\newblock {\em CoRR}, abs/1710.01878, 2017.

\end{thebibliography}
}

\end{document}